\pdfoutput=1
\documentclass[10pt]{article}
\usepackage[letterpaper,margin=1in]{geometry}
\usepackage[T1]{fontenc}
\usepackage{amsmath}
\usepackage{newpxtext}
\usepackage{newpxmath}

\usepackage{graphicx,booktabs}
\usepackage[round]{natbib}
\usepackage{microtype}
\usepackage{xcolor}
\definecolor{accent}{HTML}{00674F}
\definecolor{boxhead}{HTML}{009170}
\definecolor{boxlight}{HTML}{3EBB9E}
\usepackage[font=small,labelfont=bf,labelsep=quad]{caption}
\usepackage{titlesec}
\titleformat{\section}{\color{accent}\large\bfseries}{\thesection}{1em}{}
\titleformat{\subsection}{\color{accent}\normalsize\bfseries}{\thesubsection}{1em}{}
\titleformat{\paragraph}{\normalfont\normalsize\bfseries}{}{0pt}{}
\titlespacing*{\section}{0pt}{1.2em plus 0.3em minus 0.2em}{0.7em}
\titlespacing*{\subsection}{0pt}{1.1em plus 0.2em minus 0.1em}{0.5em}
\titlespacing*{\paragraph}{0pt}{0.7em plus 0.2em minus 0.1em}{0.1em}
\usepackage[skins]{tcolorbox}
\usepackage[colorlinks=true,linkcolor=accent,citecolor=accent,urlcolor=accent]{hyperref}
\usepackage{url}
\hypersetup{pdftitle={FACT: Fixed-Budget Allocation and Cross-Step Training for Speculative Decoding},pdfauthor={Hadi Fares, Mayssa Hajj Hassan, Ali Salman, Ammar Mohanna}}

\begin{document}
\thispagestyle{plain}

\vspace*{-0.2in}
{\color{accent}\rule{\linewidth}{1.6pt}}\par
\vspace{10pt}
{\centering
{\fontsize{17.5}{22.5}\selectfont\bfseries FACT: Fixed-Budget Allocation and Cross-Step\\
Training for Speculative Decoding\par}}
\vspace{8pt}
{\color{accent}\rule{\linewidth}{0.6pt}}\par
\vspace{14pt}
{\centering
{\large\bfseries Hadi~Fares, Mayssa~Hajj~Hassan, Ali~Salman, Ammar~Mohanna\par}
\vspace{5pt}
American University of Beirut\par}
\vspace{16pt}

{\let\thefootnote\relax\footnotetext{Correspondence to: \texttt{\{hnf01, mhh77, aas145\}@mail.aub.edu}. Code: \url{https://github.com/mrfares11/specdec}.}}

\begin{tcolorbox}[enhanced,colback=boxlight!7!white,colframe=boxlight!45!white,
  boxrule=0.6pt,arc=4mm,left=6mm,right=6mm,top=4mm,bottom=4mm]
{\centering\color{boxhead}\large\bfseries Abstract\par}
\vspace{5pt}
Speculative decoding accelerates language-model generation by verifying multiple draft tokens in a single target-model pass. However, better draft predictions and longer accepted continuations do not necessarily produce faster generation. To examine this gap, we present FACT (Fixed-Budget Allocation and Cross-Step Training), an experimental framework for studying how drafter training, candidate allocation, and execution conditions jointly affect acceptance and latency. Using a lightweight feature-prediction drafter with a frozen Llama-3-8B-Instruct target under greedy decoding, we first examine how training translates into accepted tokens. Backpropagating through predicted-feature rollouts with greater weight on early steps improves acceptance by 15--24\% over uniformly weighted detached training, yet average rollout proxy cross-entropy does not reliably rank drafters by acceptance. Beyond prediction quality, the placement of draft candidates also matters: with a fixed 256-node budget and matched per-depth widths, replacing static allocation with confidence-guided allocation increases speedup over autoregressive decoding from $1.46\times$ to $2.02\times$ on HumanEval and from $1.31\times$ to $1.75\times$ on MT-Bench. This allocation benefit extends to an independently trained drafter, but stronger training does not consistently amplify it, showing that the two improvements do not necessarily reinforce each other. Even these speed gains depend on execution conditions: changing only host-thread settings can nearly eliminate speedup without changing acceptance. Together, these findings show why training and allocation must be evaluated jointly using both acceptance and end-to-end latency under controlled execution conditions.
\end{tcolorbox}

\vspace{4pt}
\section{Introduction}
\label{sec:introduction}

Autoregressive language models generate text one token at a time, requiring a new model evaluation for each successive token. Speculative decoding reduces this sequential cost: a lightweight draft model proposes several candidate tokens, and the larger target model checks them in parallel. Generation can then advance by multiple tokens per target-model call while preserving the target's output distribution under exact arithmetic \citep{leviathan2023speculative,chen2023accelerating}.

Realizing this promise requires useful proposals at a sufficiently low cost. EAGLE improves drafting by reusing the target model's internal features \citep{li2024eagle}, while DistillSpec and HASS improve alignment with the target's predictions and the inputs encountered during decoding \citep{zhou2024distillspec,zhang2024hass}. Because the next token can remain uncertain, tree-based methods explore several continuations in parallel \citep{miao2024specinfer}. Building on this flexibility, EAGLE-2 adapts candidate allocation to the context \citep{li2024eagle2}, while Sequoia also considers hardware when selecting tree size and depth \citep{chen2024sequoia}.

Yet lower prediction loss, longer accepted continuations, and faster generation remain different outcomes. Average loss can obscure errors at early drafting steps, even though later tokens are useful only when the preceding continuation is accepted. Similarly, a tree that covers more useful continuations may take longer to construct and process. Improvements therefore depend on both the tokens gained and the time spent obtaining them.

These distinctions motivate three research questions. RQ1. Training: How do rollout training choices affect acceptance, and does held-out prediction loss reflect those changes? RQ2. Allocation: Under fixed candidate counts at each depth, how does confidence-guided allocation affect acceptance and latency across different drafters? RQ3. Execution: How do decode-loop implementation and host-thread settings affect whether acceptance gains translate into speed?

\begin{figure}[htbp]
\centering
\includegraphics[width=\linewidth]{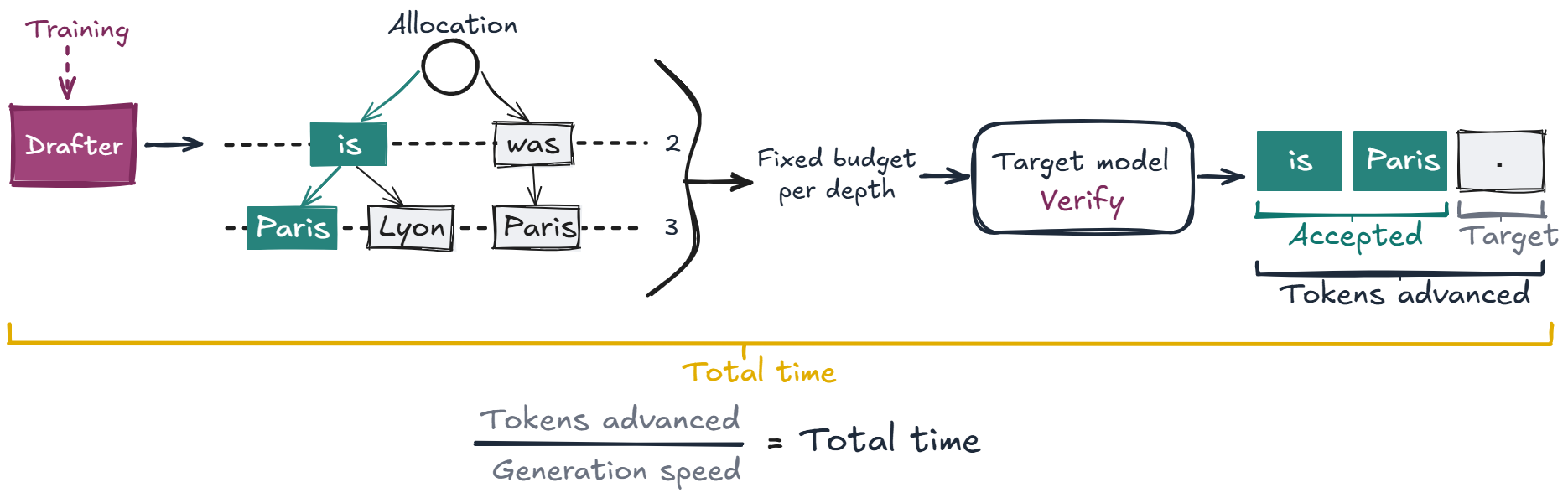}
\caption{From draft proposals to generation speed. A lightweight drafter proposes candidate continuations, which are arranged in a tree and checked by the target model. Accepted draft tokens and a target-generated token advance the output. We study drafter training, candidate allocation under fixed depth-wise budgets, and execution conditions, measuring both useful progress and total decoding time.}
\label{fig:intro-overview}
\end{figure}

We organize these comparisons in FACT (Fixed-Budget Allocation and Cross-Step Training), a shared experimental framework built around an EAGLE-style drafter and a frozen Llama-3-8B-Instruct target. FACT combines established drafting, rollout-training, and cache-reuse ideas to isolate their empirical effects; the contribution is the controlled comparison and its findings. Our comparisons vary training, allocation, and execution settings, and test whether allocation gains transfer to an independently trained official EAGLE drafter. As shown in Figure~\ref{fig:intro-overview}, we connect these choices by measuring both useful progress and total decoding time.

Our contributions are threefold:
\begin{itemize}
\item C1. Rollout training improves acceptance, but proxy loss can misrank drafters. Step-decayed, full-gradient rollout improves acceptance by 15--24\% over the tested detached baseline in eight comparisons spanning two benchmarks, two decoding depths, and two training seeds. Yet a teacher-forced drafter can outperform detached rollout in acceptance despite substantially worse multi-step proxy cross-entropy.
\item C2. Allocation gains transfer under matched depth-wise budgets. Confidence-guided allocation reduces latency without increasing candidate counts at any depth. This benefit transfers to the official EAGLE drafter, with paired latency improvements whose confidence intervals exclude zero in all ten evaluated benchmark--horizon combinations. However, stronger drafter training does not consistently increase the benefit of dynamic allocation.
\item C3. Execution settings can nearly eliminate speedup without changing acceptance. Changing only the host-thread configuration moves dynamic decoding from $1.99\times$ to $1.03\times$ speedup on HumanEval, with unchanged acceptance. Instrumented runs locate the additional cost in tree construction and the target forward pass.
\end{itemize}

\section{Background and Related Work}
\label{sec:background}

\paragraph{Speculative decoding}
\label{sec:background:specdec}
Speculative sampling verifies draft proposals using target probabilities and corrects rejected proposals to preserve the target distribution under exact arithmetic \citep{leviathan2023speculative,chen2023accelerating}. We evaluate greedy decoding: the verifier retains the longest candidate path matching the target's argmax choices and appends a target-generated correction or bonus token. Accepting $k$ draft tokens therefore advances generation by $k+1$ tokens, except when an end-of-sequence token or output budget terminates the iteration. We report accepted draft tokens separately from total emitted tokens. Appendix~\ref{app:decoding-rule} gives the sampling rule for context; the reported acceleration results concern greedy decoding.

\paragraph{Feature-based drafting}
\label{sec:background:eagle}
EAGLE predicts target-model features using a lightweight drafter conditioned on preceding features and a token sequence shifted one position ahead \citep{li2024eagle}. Its objective combines feature regression with a token-distribution loss through the frozen target readout. Noise added to training features improves robustness but does not explicitly reproduce a drafter's own feature rollouts. We retain this feature-prediction architecture while studying different rollout objectives. Our feature extraction and readout convention differs from that of the official checkpoint; Section~\ref{sec:method:framework} makes that distinction explicit.

\paragraph{Candidate allocation}
\label{sec:background:trees}
Tree attention lets candidate nodes attend to the shared prefix and their ancestors, so the target can score several continuations in one call \citep{miao2024specinfer}. EAGLE-2 uses cumulative draft confidence to expand promising paths and then reranks candidates across depths \citep{li2024eagle2}; Sequoia considers hardware when selecting tree size and depth \citep{chen2024sequoia}. Our comparison adopts cumulative-confidence selection but fixes the candidate count at every depth and omits final reranking across depths. It isolates allocation among parents under matched depth budgets, rather than comparing complete native decoding systems.

\paragraph{Training for multi-step drafting}
\label{sec:background:training}
DistillSpec improves draft--target alignment through distillation \citep{zhou2024distillspec}. HASS trains with reused predicted features, introduces a distillation loss focused on high-probability target tokens, and investigates geometrically decaying weights on later training steps \citep{zhang2024hass}. EAGLE-3 incorporates training-time drafting, removes feature reconstruction, and fuses features from multiple target layers \citep{li2025eagle3}. We do not claim rollout exposure or geometric step weighting as new mechanisms. We compare teacher forcing, uniformly weighted detached feature rollouts, and step-decayed full-gradient rollouts under a shared feature/readout convention. Our detached arm is a controlled training baseline, not a reproduction of the complete HASS system or objective. The replicated comparison tests the combined change in weighting and gradient flow; a separate exploratory control is reported in Appendix~\ref{app:training-control}.

\section{Methodology}
\label{sec:methodology}

FACT varies drafter training, candidate allocation, and execution within a shared speculative decoding framework. This lets us examine how proposal quality and allocation affect useful progress, while accounting for the cost of producing and verifying candidates.

\subsection{Shared drafting framework}
\label{sec:method:framework}

We use a trainable fusion layer and a single decoder layer with frozen target embeddings and readout, following the EAGLE architecture \citep{li2024eagle}. Let $f_i$ denote the target feature associated with recorded token $x_i$. For our trained drafters, this is the input to the target's last decoder layer. The readout is
\begin{equation}
H(f)=\operatorname{softmax}\!\left(W_{\mathrm{LM}}\operatorname{Norm}(f)\right).
\label{eq:proxy-readout}
\end{equation}
Because $f_i$ precedes the last decoder layer, $H(f_i)$ is a \emph{proxy distribution}, not the full target's next-token distribution $p(\cdot\mid x_{1:i})$. The training loss matches predictions to this proxy; verification always uses the full target. We preserve this feature/readout convention across our training arms, making their comparison conditional on this design. The independently trained official EAGLE checkpoint uses its own feature convention, specified in Appendix~\ref{app:method-details}. Training rollout length $R$ and maximum decoding depth $K$ are independent choices.

\subsection{Multi-step drafter training}
\label{sec:method:training}

During drafting, predicted features become inputs to later predictions. Following the feature-alignment motivation of HASS \citep{zhang2024hass}, we expose the drafter to these imperfect inputs during training. The first rollout step pairs recorded tokens with preceding target features. Later steps reuse predicted features, shifting the token and supervision sequences accordingly. Conditioning tokens remain those recorded in the training data; only the feature inputs are replaced by predictions.

At each step $s$, we use a combined feature and proxy-token objective: SmoothL1 loss $\mathcal L_{\mathrm{feat}}^{(s)}$, averaged over feature dimensions and valid positions, and proxy cross-entropy $\mathcal L_{\mathrm{tok}}^{(s)}=\operatorname{CE}(H(f),H(\hat f))$, averaged over valid positions. We combine them as
\begin{equation}
\mathcal L_{\mathrm{rollout}}
=
\sum_{s=1}^{R} w_s\,\mathcal L_{\mathrm{feat}}^{(s)}
+
\lambda\sum_{s=1}^{R} w_s\,\mathcal L_{\mathrm{tok}}^{(s)},
\qquad
w_s=\frac{\gamma^{s-1}}{\sum_{r=1}^{R}\gamma^{r-1}},
\label{eq:rollout-objective}
\end{equation}
where $\lambda=0.1$. Uniform weighting uses $\gamma=1$, while step-decayed weighting uses $\gamma=0.5$. In both cases, the step weights sum to one; this normalizes their total coefficient, not the resulting loss values or gradient magnitudes. Earlier steps receive more weight because accepting later tokens depends on accepting their preceding continuation. HASS also investigates this motivation \citep{zhang2024hass}.

\begin{figure}[htbp]
\centering
\includegraphics[width=0.8\linewidth]{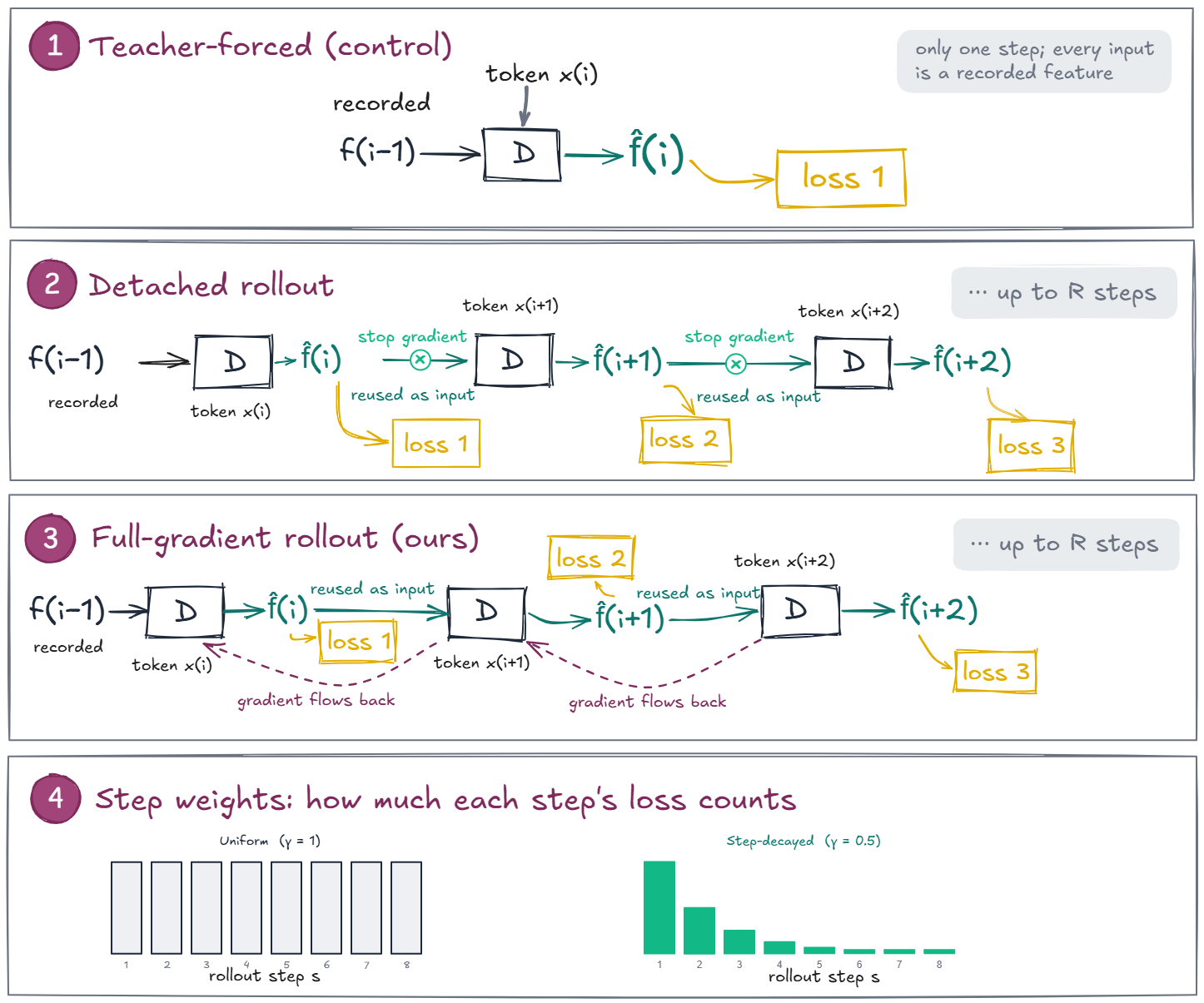}
\caption{Training on predicted features. The drafter $D$ predicts the next feature from a recorded feature and token. \textcircled{1} Teacher forcing trains one step on recorded features only. \textcircled{2} Detached rollout reuses each prediction as the next input but stops gradients between steps. \textcircled{3} Full-gradient rollout keeps the reuse and lets later losses reach earlier steps. \textcircled{4} Step weights control each prediction's contribution to the loss: uniform ($\gamma=1$) or step-decayed ($\gamma=0.5$).}
\label{fig:training-rollout}
\end{figure}

Weighting and gradient flow are separate choices. Detached rollout stops gradients through features reused at the next step; full-gradient rollout lets later losses influence earlier predictions. The replicated comparison uses uniform detached rollout, step-decayed full-gradient rollout, and a one-step teacher-forced control using target input features. An exploratory uniform full-gradient control was evaluated at one seed on an earlier prompt set (Appendix~\ref{app:training-control}). Figure~\ref{fig:training-rollout} illustrates these distinctions.

\subsection{Candidate allocation under fixed budgets}
\label{sec:method:allocation}

Given the drafter's predictions, allocation determines which continuations receive candidate slots. We fix a depth-wise budget $(b_1,\ldots,b_K)$ and total candidate count $B=\sum_{d=1}^{K}b_d$. Both policies use the same counts. The static policy gives parents predetermined child quotas, filled with each parent's highest-probability tokens. Token identities depend on the context, but the branching structure does not.

The dynamic policy lets extensions from all retained parents compete for the slots at each depth. It scores a continuation $y_{1:d}$ by cumulative draft confidence:
\begin{equation}
S(y_{1:d})
=
\sum_{j=1}^{d}
\log q\!\left(y_j\mid x_{1:i},y_{1:j-1}\right),
\label{eq:candidate-score}
\end{equation}
where $x_{1:i}$ is the current prefix and $q$ the draft distribution. At depth $d$, it retains the $b_d$ highest-scoring extensions of retained depth-$(d-1)$ parents. One parent may therefore receive several slots and another none.

This follows EAGLE-2's cumulative-confidence principle \citep{li2024eagle2}, but fixes the budget at every depth and omits its final reranking across depths. Matching candidate counts and depths does not imply equal execution cost: dynamic selection and tree construction remain part of decoding time. Appendix~\ref{app:method-details} specifies the quota rule and gives an illustrative allocation example.

\subsection{Verification and cache reuse}
\label{sec:method:verification}

The target checks the tree using an attention mask that exposes the shared prefix and each node's ancestors, excluding competing branches. Greedy verification retains the path matching the target's highest-probability choices and supplies a correction or bonus token, as described in Section~\ref{sec:background:specdec}. We gather the accepted branch's target key--value cache entries from this forward pass instead of recomputing them.

The target-generated token is then pending: its identity is known, but its cache entry is not yet computed. We process it with the next candidate tree in one target forward pass, using a predicted feature for its position until that pass provides the true target feature. After prompt prefill, this folded procedure uses one target forward pass per decoding iteration over the pending token and $B$ candidates.

Our dynamic allocation supports greedy decoding. Verification preserves the target's greedy continuation under exact arithmetic; finite-precision agreement is evaluated separately. Host-thread settings can change execution cost without changing the intended acceptance rule. Additional implementation details are given in Appendix~\ref{app:method-details}.

\section{Experiments}
\label{sec:experiments}

We evaluate whether improvements in speculative decoding produce both more useful draft tokens and faster generation. We first examine how training choices affect acceptance and whether prediction loss captures these differences (RQ1). We then test confidence-guided allocation under fixed depth-wise budgets and examine whether its benefits transfer across drafters (RQ2). Finally, we investigate how execution conditions affect the conversion of accepted tokens into generation speed (RQ3).

\subsection{Experimental setup}
\label{sec:experiments:setup}

We use a frozen Llama-3-8B-Instruct target \citep{dubey2024llama3} and evaluate greedy decoding on a single NVIDIA H100 in bfloat16. Our workloads comprise prompts from MT-Bench \citep{zheng2023mtbench}, HumanEval \citep{chen2021humaneval}, MBPP \citep{austin2021mbpp}, GSM8K \citep{cobbe2021gsm8k}, and Alpaca \citep{taori2023alpaca}. Main comparisons use 50 held-out test prompts per benchmark; the host-thread sweep uses separate development prompts. We use 256 candidate nodes, decoding depths $K\in\{4,8\}$, and a 128-token generation budget. Host threads are capped at eight unless varied explicitly.

We report accepted draft tokens per verification step, excluding the target-generated token, together with latency in milliseconds per generated token and speedup over autoregressive decoding. Measurements include prefill but exclude model loading, and latency is normalized by the actual number of emitted tokens. Repeated measurements are averaged within each prompt before computing paired comparisons; prompts then receive equal weight. Allocation timings use three to ten repeats per configuration, with the main $K=4$ comparisons on MT-Bench and HumanEval spanning three sessions. Confidence intervals use 10,000 paired bootstrap resamples over prompts. Table~\ref{tab:protocol} consolidates the reported settings. Speedups are paired summaries, so they need not equal ratios reconstructed from rounded mean latencies.

\subsection{RQ1: How does training affect acceptance?}
\label{sec:experiments:training}

We compare teacher forcing, uniformly weighted detached rollout, and step-weighted backpropagation through the rollout with decay $\gamma=0.5$. The three objectives use the same 80,000 feature rows, 10,000 optimization steps, and batch configuration. These controls match data and optimizer steps, not training compute. Rollout-trained models use $R=8$ training steps, while teacher forcing uses one. We evaluate two independently trained seeds on the same static tree, separating training effects from candidate allocation.

\begin{figure}[htbp]
\centering
\includegraphics[width=\linewidth]{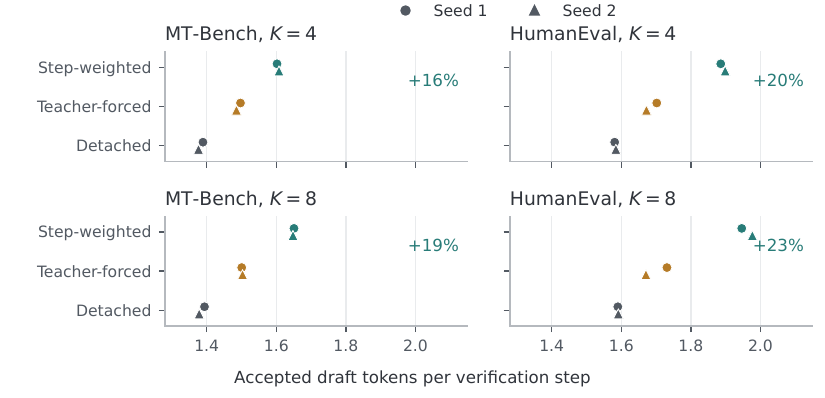}
\caption{Training improvements persist across seeds. Each marker shows mean draft acceptance over 50 test prompts for one training seed. All configurations use a static 256-node tree. Percentages compare step-weighted training with detached rollout after averaging the two seeds. $K$ denotes decoding depth, not training rollout length.}
\label{fig:exp-training}
\end{figure}

As shown in Figure~\ref{fig:exp-training}, step-weighted training improves acceptance over detached rollout in all eight combinations of benchmark, decoding depth, and seed, with gains ranging from 15\% to 24\%. At $K=4$, acceptance averaged across seeds increases from 1.38 to 1.60 on MT-Bench and from 1.58 to 1.89 on HumanEval. These comparisons support the combined choice of step weighting and cross-step backpropagation; they do not separately identify the contribution of each component.

\begin{figure}[htbp]
\centering
\includegraphics[width=\linewidth]{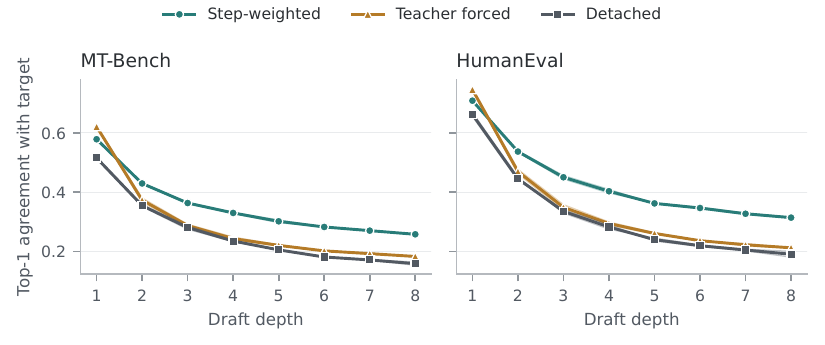}
\caption{Agreement with the target at each draft depth. Fraction of positions where the drafter's top-ranked token equals the target's greedy next token, measured along the target's continuation of 50 test prompts per benchmark with decode-time feature inputs. Thick lines average training seeds 1 and 2, and thin lines show each seed. The checkpoints are the 10,000-step drafters of Figure~\ref{fig:exp-training}.}
\label{fig:exp-depth}
\end{figure}

Figure~\ref{fig:exp-depth} shows where these gains arise. We roll each drafter out along the target's greedy continuation of each test prompt, in the same way as a chain of the draft tree at decode time, and record how often its top-ranked token matches the target's next token at each draft depth. Teacher forcing agrees most often at the first depth, but its agreement falls fastest after that. From the second depth onward, step-weighted training agrees most often on both benchmarks, while detached rollout agrees least at every depth.

However, held-out rollout proxy cross-entropy does not reproduce this ordering. Under a common proxy cross-entropy evaluator that weights all eight rollout steps equally, teacher forcing scores substantially worse than detached training, 7.86 versus 3.73, yet accepts more tokens on MT-Bench at $K=4$, 1.49 versus 1.38. Meanwhile, detached and step-weighted training have similar proxy cross-entropy, 3.73 and 3.71, despite their different acceptance. Thus, this proxy metric does not consistently rank the tested objectives by acceptance. This finding is conditional on the intermediate-feature readout and does not establish that cross-entropy against the full target distribution is generally uninformative.

\subsection{RQ2: Does candidate allocation improve generation across drafters?}
\label{sec:experiments:allocation}

Having examined training under a fixed tree, we next hold drafter weights fixed and change how candidates are allocated. Static and dynamic trees use identical numbers of nodes at every depth. The comparison therefore changes where candidate slots are assigned without increasing the candidate budget, although tree-construction costs can still differ.

We evaluate an earlier baseline drafter, our scaled drafter, and the independently trained official EAGLE drafter \citep{li2024eagle}. The scaled checkpoint was trained for 20,000 steps and is distinct from the 10,000-step checkpoints used in RQ1.

Dynamic allocation improves both acceptance and generation time. For the scaled drafter at $K=4$, MT-Bench acceptance rises from 1.92 to 2.66 while latency falls from 13.96 to 10.32 ms/token, increasing speedup from $1.31\times$ to $1.75\times$. On HumanEval, acceptance rises from 2.27 to 3.22 and latency falls from 12.47 to 8.93 ms/token, increasing speedup from $1.46\times$ to $2.02\times$.

\begin{table}[t]
\centering
\caption{Main allocation comparison. Scaled drafter, $K=4$, 256 candidates, and 50 test prompts per workload. $\tau$ excludes the target-generated token; latency includes prefill. Figure~\ref{fig:exp-allocation} reports uncertainty on paired latency savings.}
\label{tab:main-results}
\small
\begin{tabular}{llrrr}
\toprule
Workload & Allocation & $\tau$ & ms/token & Speedup \\
\midrule
MT-Bench & Static & 1.92 & 13.96 & $1.31\times$ \\
MT-Bench & Dynamic & 2.66 & 10.32 & $1.75\times$ \\
HumanEval & Static & 2.27 & 12.47 & $1.46\times$ \\
HumanEval & Dynamic & 3.22 & 8.93 & $2.02\times$ \\
\bottomrule
\end{tabular}
\end{table}

\begin{figure}[htbp]
\centering
\includegraphics[width=\linewidth]{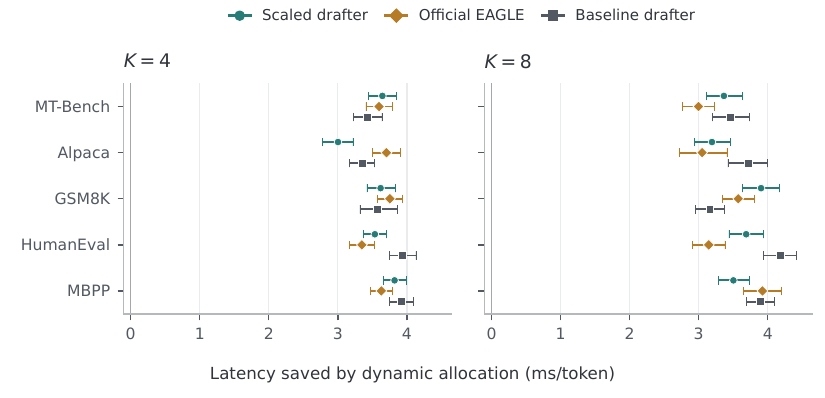}
\caption{Allocation gains transfer across drafters. Points show paired latency savings, $L_{\mathrm{static}}-L_{\mathrm{dynamic}}$, under identical depth-wise candidate budgets. Positive values favor dynamic allocation. Bars show 95\% bootstrap intervals over 50 test prompts after pooling repetitions within each prompt. Both panels use the same horizontal scale.}
\label{fig:exp-allocation}
\end{figure}

Figure~\ref{fig:exp-allocation} extends this comparison across all five workloads. Applied unchanged to the official EAGLE drafter, dynamic allocation saves 3.00--3.92 ms/token, with confidence intervals excluding zero in all ten benchmark--depth combinations. This demonstrates transfer across the evaluated drafters under a shared target and decoding implementation; it is not a comparison against the native EAGLE-2 system.

\begin{figure}[htbp]
\centering
\includegraphics[width=\linewidth]{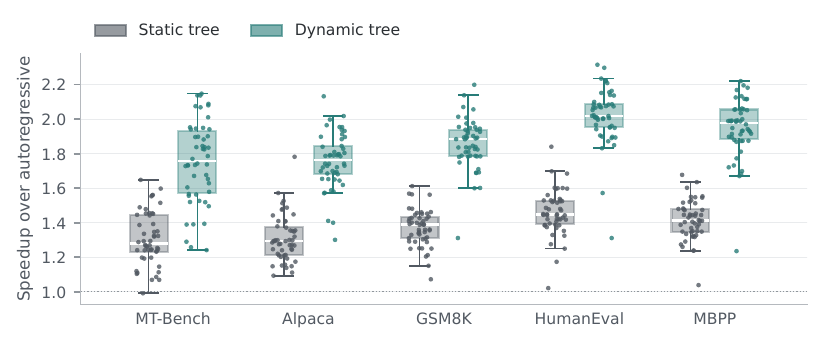}
\caption{Per-prompt speedups for the scaled drafter at $K=4$. Each point is one of the 50 test prompts per workload, with repeated runs averaged within the prompt; HumanEval and MT-Bench pool ten runs and the other workloads three. Boxes show quartiles and medians. Speedup is the per-prompt ratio to autoregressive decoding from the same run, and the dotted line marks autoregressive speed.}
\label{fig:exp-prompts}
\end{figure}

Figure~\ref{fig:exp-prompts} shows the same comparison prompt by prompt for the scaled drafter at $K=4$. Dynamic allocation is faster than static allocation on every one of the 250 prompts across the five workloads, so the mean gains in Table~\ref{tab:main-results} do not come from a few favorable prompts.

We measure interaction on the absolute latency-saving scale. For drafter $d$, define $G(d)=L_{\mathrm{static}}(d)-L_{\mathrm{dynamic}}(d)$ and
\begin{equation}
I=G(\mathrm{scaled})-G(\mathrm{baseline}).
\label{eq:interaction}
\end{equation}
Positive $I$ means that allocation saves more milliseconds per token for the scaled drafter. The estimate is negative with an interval excluding zero in five of ten benchmark--depth combinations, positive only on GSM8K at $K=8$, and inconclusive in four. For HumanEval at $K=4$, $I=-0.40$ ms/token with a 95\% interval of $[-0.64,-0.17]$ (Table~\ref{tab:interaction}). Thus, a stronger drafter does not consistently receive a larger absolute latency reduction from allocation. This statement concerns the evaluated checkpoints and this interaction scale; it does not imply an identical interaction on a speedup-ratio scale.

\subsection{RQ3: When do acceptance gains translate into speed?}
\label{sec:experiments:execution}

The preceding results use a fixed execution configuration. To test how much this matters, we vary the host-thread cap while keeping the code, GPU, drafter weights, prompts, and candidate budget unchanged.

\begin{figure}[htbp]
\centering
\includegraphics[width=\linewidth]{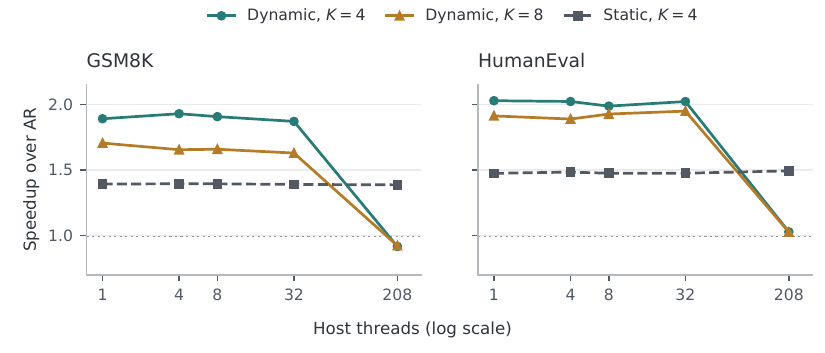}
\caption{Execution conditions change speed without changing acceptance. Speedup over autoregressive decoding as host-thread count varies. Each point averages paired speedup ratios over 50 development prompts from one run per setting. The dotted line denotes autoregressive speed. Acceptance remains unchanged across caps within each configuration.}
\label{fig:exp-host-threads}
\end{figure}

Figure~\ref{fig:exp-host-threads} shows that dynamic decoding remains effective at caps between one and 32 threads but loses most of its advantage at 208 threads. At $K=4$, moving from eight to 208 threads reduces speedup from $1.91\times$ to $0.92\times$ on GSM8K and from $1.99\times$ to $1.03\times$ on HumanEval. The $K=8$ dynamic configuration shows the same pattern, whereas the static tree remains comparatively stable.

Crucially, draft acceptance is identical across thread settings within each configuration. The change in speed therefore occurs without a change in accepted draft progress.

\begin{figure}[htbp]
\centering
\includegraphics[width=\linewidth]{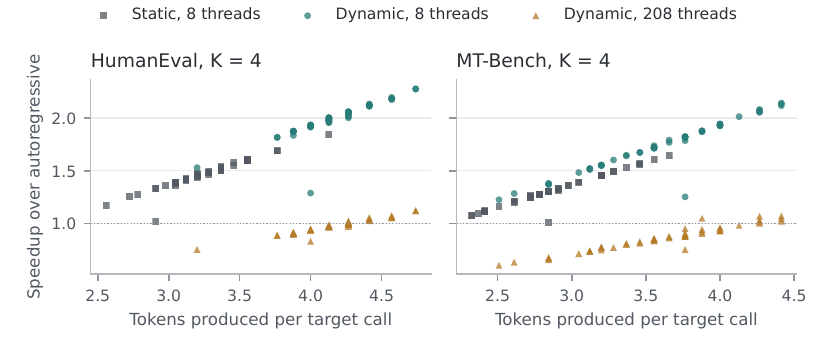}
\caption{Tokens per target call and speedup for each prompt. Each point is one test prompt averaged over five runs per setting, using the scaled drafter. Tokens per target call include the target-generated token. The static tree at 208 threads, not shown, matches the static tree at eight threads. The dotted line marks autoregressive speed.}
\label{fig:exp-accept-speed}
\end{figure}

A separate campaign repeats the comparison five times on the 50 test prompts of HumanEval and MT-Bench at $K=4$. Figure~\ref{fig:exp-accept-speed} plots each prompt's tokens per target call against its speedup. At 208 threads every prompt produces exactly the same number of tokens per target call as at eight threads, yet the mean speedup of dynamic decoding falls from $1.99\times$ to $0.98\times$ on HumanEval and from $1.74\times$ to $0.86\times$ on MT-Bench. Static decoding keeps its speedup at both settings.

Separate instrumented HumanEval runs help locate the added cost. Increasing the thread cap from eight to 208 raises tree-construction time from 3.19 to 6.40 ms per step and target tree-forward time from 20.83 to 51.94 ms, while acceptance checking remains near 0.15 ms. These phase timings are diagnostic measurements rather than the uninstrumented timings used for headline speedups. They show why inexpensive acceptance checking alone is insufficient: execution overhead elsewhere in the loop can erase the benefit of higher acceptance.

\begin{figure}[htbp]
\centering
\includegraphics[width=\linewidth]{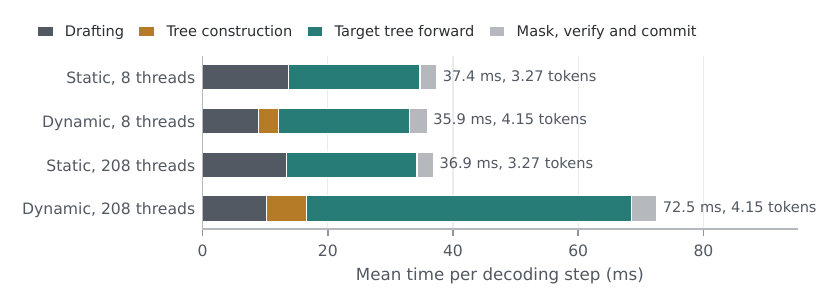}
\caption{Time per decoding step by phase. Means over 25 HumanEval development prompts with the scaled drafter at $K=4$ and phase timers enabled, weighting each prompt by its number of steps. Labels give the total time and the tokens emitted per step. Timers can perturb execution, so these runs are diagnostic and are not used for the speedups reported elsewhere.}
\label{fig:exp-cost}
\end{figure}

Figure~\ref{fig:exp-cost} shows the complete per-step breakdown from these runs. The static tree costs about the same per step at both settings, 37.4 and 36.9 ms. For the dynamic tree, a step grows from 35.9 to 72.5 ms, and the target tree forward pass and tree construction account for 34.3 ms of the 36.6 ms increase.

\paragraph{Numerical fidelity}
We assess output agreement separately from speed. Float32 decoding matches the autoregressive target at all 52,224 tested positions. In bfloat16, same-prefix agreement is 99.0\%, so the speed measurements should not be interpreted as guaranteeing identical token sequences. Separate quality probes on 60 prompts per benchmark yield paired differences of $+0.033$ on HumanEval pass@1 (95\% interval $[-0.033,0.100]$) and $0.000$ on GSM8K exact match ($[-0.067,0.067]$). These intervals do not establish quality equivalence. Appendix~\ref{app:numerical} describes these probes and distinguishes them from the 50-prompt timing evaluations.

\section{Discussion and Limitations}
\label{sec:discussion-limitations}

\paragraph{Acceptance gains are not output-rate gains}
Our acceptance metric excludes the target-generated token. Away from termination and output-cap effects, increasing mean draft acceptance from $\tau_0$ to $\tau_1$ increases emitted tokens per iteration by a factor $(\tau_1+1)/(\tau_0+1)$, before accounting for iteration cost. For the RQ1 MT-Bench means, the increase from 1.38 to 1.60 is about 16\% in draft acceptance but only 9.2\% in tokens advanced per iteration. Prefill and changes in drafting or target-forward cost further affect measured latency. This explains why acceptance percentages should not be read as expected speedup percentages.

\paragraph{Proxy loss and model selection}
The RQ1 result concerns ranking different training objectives using the proxy cross-entropy defined by our intermediate-feature readout. That readout does not produce the full target distribution, so proxy--target mismatch is one possible contributor to the observed ranking failure. We do not isolate it from other effects such as how errors are distributed over rollout steps. The result also does not rule out useful within-run correlations between loss and acceptance. Its supported implication is to evaluate acceptance directly when choosing among these objectives.

\paragraph{Allocation and execution}
Matched depth budgets isolate changes in candidate placement while keeping dynamic selection costs inside the measurement. The positive latency savings across the evaluated checkpoints show that the allocation benefit is not restricted to our trained weights. The interaction remains scale-specific: stronger training does not consistently increase absolute milliseconds saved. The thread sweep further shows that an unchanged acceptance count can accompany a large latency change. These results motivate recording execution settings alongside the allocation policy and checkpoint.

\paragraph{Scope and attribution}
The evaluation uses one target model, an H100 GPU platform, short generation budgets, and 50 test prompts per benchmark. It does not establish acceleration for sampling, longer outputs, or batched serving, or superiority over native EAGLE-2. The replicated training comparison uses two benchmarks and two seeds and changes both weighting and gradient flow; it supports their combined configuration. The uniform full-gradient control is exploratory and uses an earlier prompt set. The scaled allocation checkpoint has 20,000 training steps, whereas the RQ1 checkpoints have 10,000, so the allocation results do not directly measure the latency effect of the replicated RQ1 intervention.

\paragraph{Measurement and numerical limits}
Prompt-level bootstrap intervals quantify variation across the evaluated prompts, not all session or deployment variability. The thread sweep uses one run per setting, and instrumented phase timings can perturb execution. Measurements include prefill and use a 128-token output cap, so their ratios depend on workload lengths as well as decoding. Exact-arithmetic verification does not guarantee identical bfloat16 outputs. The separate quality probes remain compatible with losses of 3.3 percentage points on HumanEval and 6.7 points on GSM8K; they establish neither equivalence nor broad quality preservation.

\section{Conclusion}
\label{sec:conclusion}

We studied how drafter training, candidate allocation, and execution conditions jointly affect speculative decoding. Step-weighted full-gradient rollout improves acceptance over the tested detached baseline, while rollout proxy cross-entropy can misrank the tested objectives by acceptance. Confidence-guided allocation improves generation under fixed depth-wise budgets and transfers to an independently trained drafter, but its benefit does not consistently increase with drafter strength. Finally, execution settings can erase speed gains without changing acceptance. The central lesson is to evaluate speculative decoding through both useful progress and total decoding time, with numerical fidelity assessed separately.

\bibliography{references}
\bibliographystyle{plainnat}

\appendix
\section{Decoding rule and token accounting}
\label{app:decoding-rule}

The sampling rule is included as background, not as an acceleration result for our dynamic implementation. Let $p$ and $q$ denote target and draft distributions conditioned on an accepted prefix $x_{1:i}$. A proposed token $y\sim q$ is accepted with probability
\begin{equation}
\alpha(y\mid x_{1:i})=\min\!\left(1,\frac{p(y\mid x_{1:i})}{q(y\mid x_{1:i})}\right).
\end{equation}
On rejection, the replacement distribution is
\begin{equation}
r(y\mid x_{1:i})=
\frac{[p(y\mid x_{1:i})-q(y\mid x_{1:i})]_+}
{\sum_{v\in\mathcal V}[p(v\mid x_{1:i})-q(v\mid x_{1:i})]_+},
\end{equation}
where $[a]_+=\max(0,a)$. The residual is needed only when rejection occurs; when $p=q$, rejection has probability zero. If all proposals are accepted, a bonus token is drawn from the target. This rule preserves the target distribution under exact arithmetic \citep{leviathan2023speculative,chen2023accelerating}.

For the greedy experiments, the target supplies a correction at the first rejected position or a bonus after full acceptance. We count only matching draft tokens in $\tau$, and all emitted tokens in latency's denominator. A completed iteration normally emits $k+1$ tokens for $k$ accepted draft tokens; termination can shorten it. Algorithmic equivalence assumes consistent target logits and tie-breaking between the tree and autoregressive evaluations. Finite-precision agreement is a separate empirical check.

\section{Implementation and reported protocol}
\label{app:method-details}

\paragraph{Feature convention}
Our trained drafters use the input to the target's last decoder layer. The frozen final normalization and language-model head define their proxy distribution. The independently trained official EAGLE checkpoint uses post-normalization target features with the output head and no additional normalization. The full target model performs verification in both cases.

\paragraph{Memory-efficient training}
Full-vocabulary cross-entropy is computed in chunks of sequence positions and its gradients are accumulated through the frozen readout. This changes neither the mathematical objective nor its gradient, apart from floating-point effects. Recorded token inputs remain fixed during feature rollouts; this is exposure to predicted features, not fully generated token trajectories.

\begin{figure}[htbp]
\centering
\includegraphics[width=\linewidth]{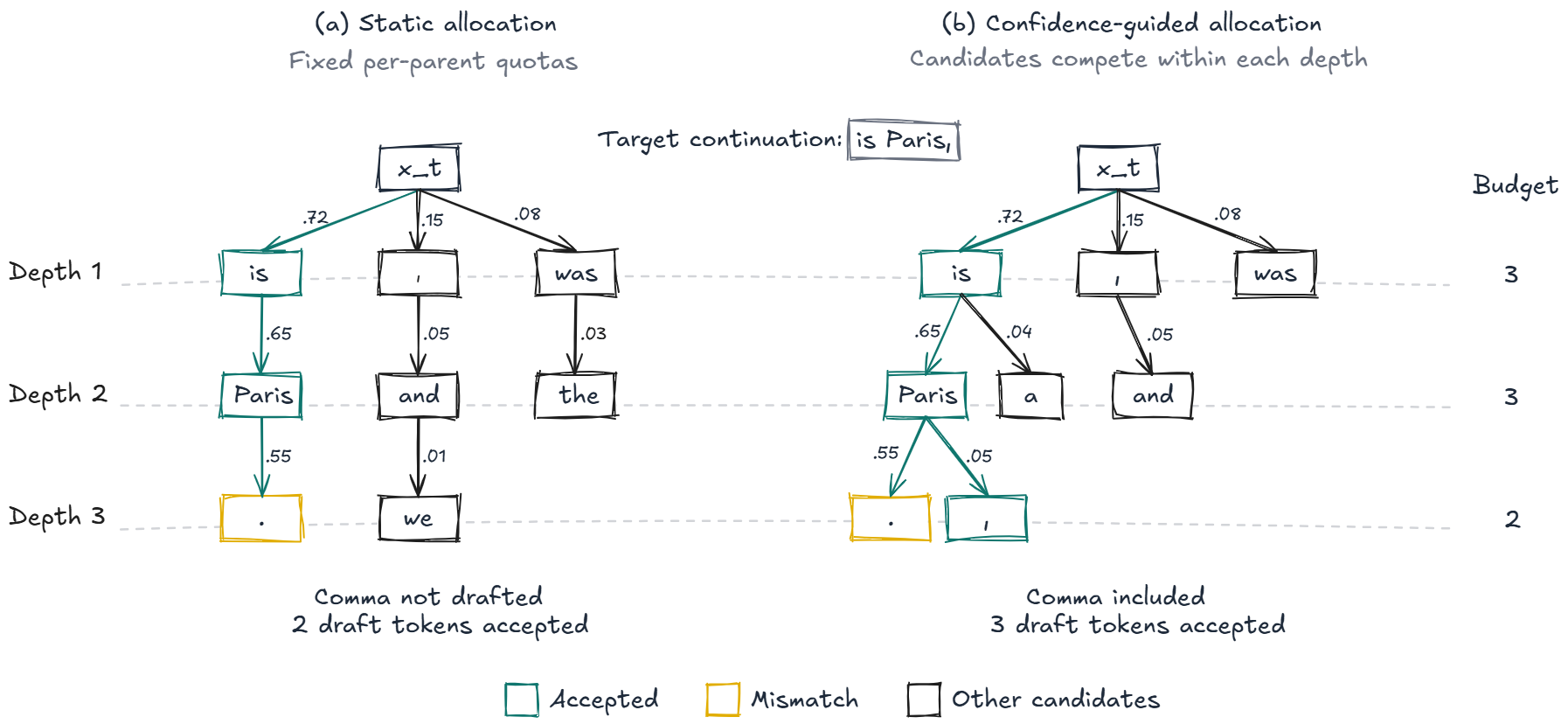}
\caption{Fixed budgets, different allocation. Illustrative widths $(3,3,2)$ and draft probabilities for the target continuation ``is Paris,''. (a) Static quotas omit the target's comma continuation, so two draft tokens are accepted before the mismatch; (b) confidence-guided (dynamic) selection places both depth-3 slots below ``Paris'' and accepts three. Accepted counts exclude the target-generated correction or bonus token.}
\label{fig:method-tree-allocation}
\end{figure}

\paragraph{Static allocation}
At depth $d$, the $b_d$ slots are divided as evenly as possible among the retained parents. A fixed parent ordering determines which parents receive any remainder. Each parent fills its quota with its highest-probability tokens. Dynamic allocation instead ranks extensions from all retained parents by cumulative draft confidence. Figure~\ref{fig:method-tree-allocation} illustrates the distinction with eight candidates; it does not specify the 256-node experimental depth profile.

\paragraph{Acceptance checking and cache commitment}
Target argmaxes are computed on the device, and compact token IDs are transferred for accepted-path selection on the host. The path identifies target key--value entries to gather. The target-generated token is processed with the next tree. The approximately 0.15 ms acceptance-checking measurement excludes target tree scoring and should not be read as the cost of the entire verification stage.

\begin{table}[htbp]
\centering
\caption{Reported settings. Training-step matching is not compute matching. The 60-prompt quality probes are separate from the timing experiments.}
\label{tab:protocol}
\small
\begin{tabular}{p{0.34\linewidth}p{0.58\linewidth}}
\toprule
Setting & Reported value \\
\midrule
Target and timing precision & Llama-3-8B-Instruct, frozen; bfloat16 \\
Execution platform & One NVIDIA H100; default host-thread cap 8 \\
Candidate budget and depth & 256 candidates; $K\in\{4,8\}$ \\
Timing prompts and output cap & 50 test prompts per benchmark; 128 tokens \\
Timing accounting & Includes prefill; excludes loading; actual emitted-token count \\
Timing repetition & 3--10 repeats per configuration; main $K=4$ HumanEval and MT-Bench comparisons span 3 sessions \\
Uncertainty & 10,000 paired bootstrap resamples over prompts \\
Thread sweep & 50 development prompts; one run per setting; caps 1, 4, 8, 32, 208 \\
RQ1 data and optimization & 80,000 feature rows; 10,000 optimizer steps; 2 seeds \\
RQ1 rollout and loss weights & $R=8$ for rollout, $R=1$ for teacher forcing; $\lambda=0.1$; $\gamma\in\{1,0.5\}$ \\
Scaled allocation checkpoint & 20,000 optimizer steps; distinct from RQ1 checkpoints \\
Quality probes & 60 prompts each on HumanEval and GSM8K; 384-token cap \\
\bottomrule
\end{tabular}
\end{table}

\section{Drafter--allocation interaction}
\label{app:interaction}

Table~\ref{tab:interaction} reports paired absolute latency savings and the interaction defined in Equation~\ref{eq:interaction}. Repeats are pooled within prompts. Intervals are 95\% percentile bootstrap intervals over prompts, with 10,000 resamples. These intervals do not imply that the same interaction sign must hold for reciprocal speedup ratios.

\begin{table}[htbp]
\centering
\caption{Dynamic-allocation savings $G$ and interaction $I$, in ms/token. Entries give estimates and 95\% intervals. Negative $I$ means smaller absolute savings for the scaled drafter.}
\label{tab:interaction}
\small
\setlength{\tabcolsep}{3pt}
\begin{tabular}{lrllll}
\toprule
Workload & $K$ & Baseline $G$ & Scaled $G$ & EAGLE $G$ & $I$ \\
\midrule
MT-Bench & 4 & 3.43 [3.22,3.64] & 3.65 [3.44,3.85] & 3.60 [3.41,3.79] & 0.22 [$-0.00$,0.44] \\
MT-Bench & 8 & 3.46 [3.19,3.73] & 3.37 [3.11,3.63] & 3.00 [2.76,3.23] & $-0.09$ [$-0.40$,0.20] \\
Alpaca & 4 & 3.35 [3.17,3.53] & 3.00 [2.78,3.22] & 3.70 [3.51,3.91] & $-0.35$ [$-0.58$,$-0.12$] \\
Alpaca & 8 & 3.72 [3.42,3.99] & 3.19 [2.93,3.46] & 3.05 [2.72,3.41] & $-0.52$ [$-0.82$,$-0.22$] \\
GSM8K & 4 & 3.57 [3.32,3.86] & 3.62 [3.42,3.83] & 3.75 [3.57,3.94] & 0.05 [$-0.30$,0.37] \\
GSM8K & 8 & 3.16 [2.96,3.37] & 3.90 [3.64,4.17] & 3.57 [3.34,3.81] & 0.74 [0.41,1.07] \\
HumanEval & 4 & 3.94 [3.75,4.13] & 3.54 [3.37,3.70] & 3.35 [3.16,3.54] & $-0.40$ [$-0.64$,$-0.17$] \\
HumanEval & 8 & 4.18 [3.94,4.42] & 3.69 [3.44,3.94] & 3.15 [2.91,3.38] & $-0.49$ [$-0.84$,$-0.12$] \\
MBPP & 4 & 3.92 [3.75,4.09] & 3.82 [3.66,3.99] & 3.63 [3.47,3.79] & $-0.10$ [$-0.31$,0.11] \\
MBPP & 8 & 3.89 [3.69,4.10] & 3.50 [3.28,3.73] & 3.92 [3.65,4.20] & $-0.39$ [$-0.67$,$-0.11$] \\
\bottomrule
\end{tabular}
\end{table}

\section{Exploratory training control and training cost}
\label{app:training-control}

Table~\ref{tab:training-control} reports an exploratory four-arm comparison. It uses one training seed and an earlier prompt set, so its acceptance values must not be pooled with the two-seed results in Section~\ref{sec:experiments:training}. Within this exploratory run, decayed weighting improves acceptance relative to uniform weighting when gradients cross rollout steps. Uniform full-gradient training does not improve on uniformly weighted detached training. No decayed-weight detached arm was evaluated, so the experiment does not isolate whether cross-step gradient flow is necessary under decayed weighting.

\begin{table}[htbp]
\centering
\caption{Exploratory four-arm control, seed 0, 80,000 feature rows and 10,000 optimizer steps. MT-Bench, static 256-node tree, earlier prompt set. CE is proxy cross-entropy. The eight-step CE evaluation is shared across objectives.}
\label{tab:training-control}
\small
\begin{tabular}{lrrrr}
\toprule
Objective & CE$_1$ & CE$_{1:8}$ & $\tau$, $K=4$ & $\tau$, $K=8$ \\
\midrule
Decayed weights, full gradient & 2.43 & 3.33 & 1.69 & 1.74 \\
Teacher forcing & 2.29 & 7.62 & 1.49 & 1.50 \\
Uniform, detached & 3.04 & 3.56 & 1.37 & 1.37 \\
Uniform, full gradient & 3.30 & 3.77 & 1.34 & 1.35 \\
\bottomrule
\end{tabular}
\end{table}

The recorded training times for 10,000 steps on one H100 are 2.4 h for teacher forcing, 14.1 h for detached rollout, and 27.0 h for step-decayed full-gradient rollout. These are diagnostic costs from the original comparison, not averages over the replicated seeds. They demonstrate that matched optimizer steps do not imply matched training time; the acceptance improvement should be interpreted with this cost difference in mind.

\begin{figure}[htbp]
\centering
\includegraphics[width=\linewidth]{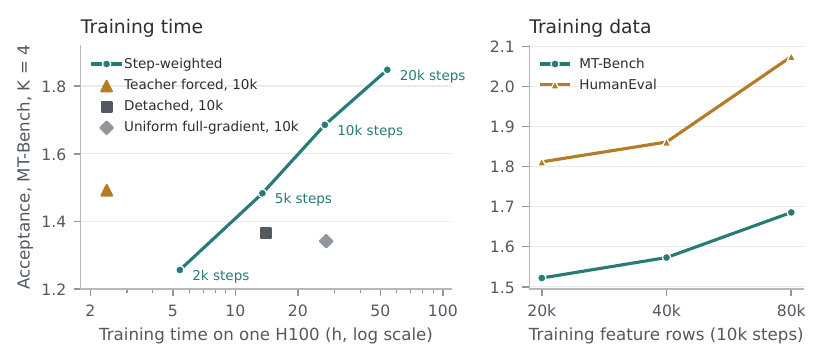}
\caption{Acceptance against training time and training data in the exploratory run (seed 0, static 256-node tree, earlier prompt set). The left panel shows MT-Bench acceptance at $K=4$. Step-weighted checkpoints are placed at the recorded rate of 27.0 h per 10,000 steps, and the other objectives at their recorded 10,000-step times, with uniform full-gradient training taken from its training log. The right panel shows step-weighted drafters trained for 10,000 steps on 20,000 to 80,000 feature rows.}
\label{fig:app-training-cost}
\end{figure}

Figure~\ref{fig:app-training-cost} places acceptance next to training time and training data in the same exploratory setting. The step-weighted drafter needs about 5,000 steps, roughly 13.5 h at its recorded rate, to reach the acceptance that teacher forcing reaches in 2.4 h, 1.48 against 1.49 on MT-Bench at $K=4$. Its acceptance is still rising at 20,000 steps, and at a fixed 10,000 steps it rises with the number of training feature rows on both benchmarks. These are single-seed results on the earlier prompt set.

\section{Numerical fidelity and quality probes}
\label{app:numerical}

Float32 speculative decoding agrees with autoregressive decoding at all 52,224 tested positions across four evaluated systems and four benchmarks. In a separate bfloat16 same-prefix comparison, 12,672 of 12,800 positions agree and 128 disagree. A free-running comparison yields 79.9\% agreement; once a token differs, the two continuations can follow different prefixes, so free-running and same-prefix agreement answer different questions.

All 128 same-prefix disagreements occur where the target's top two logits are separated by at most two bfloat16 units in the last place; approximately 92\% occur at an exact tie in the measured logits. This concentration is consistent with finite-precision sensitivity near ties. It is evidence about the observed disagreements, not a proof that every implementation path is correct. These tests do not guarantee identical bfloat16 outputs for the timed configurations.

\begin{table}[htbp]
\centering
\caption{Separate paired quality probes, 60 held-out prompts per benchmark. Differences are speculative minus autoregressive performance, on a 0--1 scale. Intervals quantify uncertainty in the difference and do not establish equivalence.}
\label{tab:quality}
\small
\begin{tabular}{llrr}
\toprule
Benchmark & Metric & Difference & 95\% interval \\
\midrule
HumanEval & pass@1 & $+0.033$ & $[-0.033,0.100]$ \\
GSM8K & Exact match & $0.000$ & $[-0.067,0.067]$ \\
\bottomrule
\end{tabular}
\end{table}

The quality probe caps generation at 384 tokens, applies no stop-sequence trimming, and extracts the last number as the GSM8K answer. These conditions are shared across the paired arms, but their effects need not cancel when continuations differ. The probe's absolute scores should not be interpreted as a standard capability evaluation. The intervals in Table~\ref{tab:quality} allow losses of 3.3 percentage points on HumanEval and 6.7 points on GSM8K. MT-Bench task quality was not evaluated with a judge model.

\end{document}